\documentclass[letterpaper, 10 pt, conference]{ieeeconf}  

\IEEEoverridecommandlockouts                              

\usepackage{graphics} 
\usepackage{amsmath} 
\usepackage{amssymb}  
\usepackage{todonotes}
\usepackage{comment}
\usepackage{diagbox}
\usepackage{makecell}
\usepackage{booktabs}
\usepackage{comment}
\usepackage{float}
\usepackage{algorithm}
\usepackage[noend]{algpseudocode}
\usepackage{varwidth}

\newtheorem{theorem}{Theorem}

\def\quad{\hskip0.5em\relax}

\DeclareMathOperator*{\argmin}{arg\,min} 

\title{\LARGE \bf
LTL-D*: Incrementally Optimal Replanning for Feasible and Infeasible Tasks in Linear Temporal Logic Specifications
}

\author{Jiming Ren, Haris Miller, Karen M. Feigh, Samuel Coogan, and Ye Zhao
\thanks{}
\thanks{
The authors are with the Institute for Robotics and Intelligent Machines, Georgia Institute of Technology, 
Atlanta, GA 30332, USA, {\tt\small jren313@gatech.edu}}
\thanks{This work is sponsored by Lockheed Martin Corporation with Steven Lincoln as Technical Monitor.   The work is that of the authors and does not represent an official position of LMCO.}
}

\begin{document}
\bstctlcite{IEEEexample: BSTcontrol} 

\maketitle
\thispagestyle{empty}
\pagestyle{empty}

\begin{abstract}
This paper presents an incremental replanning algorithm, dubbed LTL-D*, for temporal-logic-based task planning in a dynamically changing environment. Unexpected changes in the environment may lead to failures in satisfying a task specification in the form of a Linear Temporal Logic (LTL). In this study, the considered failures are categorized into two classes: (i) the desired LTL specification can be satisfied via replanning, and (ii) the desired LTL specification is infeasible to meet strictly and can only be satisfied in a ``relaxed" fashion. To address these failures, the proposed algorithm finds an optimal replanning solution that minimally violates desired task specifications. In particular, our approach leverages the D* Lite algorithm and employs a distance metric within the synthesized automaton to quantify the degree of the task violation and then replan incrementally. This ensures plan optimality and reduces planning time, especially when frequent replanning is required. Our approach is implemented in a robot navigation simulation to demonstrate a significant improvement in the computational efficiency for replanning by two orders of magnitude.

\end{abstract}

\section{Introduction}

\begin{figure}
    \centering
    \vspace{.25em}
    \includegraphics[width=\linewidth]{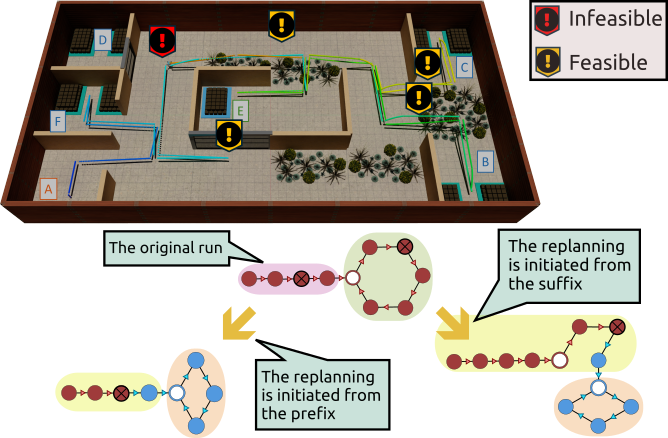}
    \caption{\textit{Top}: The trajectory of a drone starting from \textit{A} and executing the mission of carrying goods from each room \textit{F}, \textit{B}, \textit{C}, and \textit{D} to the central dropoff location \textit{E} sequentially. The color of the trajectory representing time transitions from dark blue to orange as time progresses. Multiple replanning events take place along the way where task specifications remain feasible to meet. Loading goods at \textit{D} becomes infeasible because the access to the room is closed. Therefore, the drone hovers around at its current location because all other tasks have been finished. \textit{Bottom}: We show our revision strategies based on the current phase of the run that the robot is executing. }
    \label{fig:spotlight}
    \vspace{-0.15in}
\end{figure}

As autonomous robots play an increasingly important role in handling long-horizon missions in complex environments, task and motion planning becomes essential for seamlessly integrating high-level task planning with low-level motion planning. By logically reasoning about the temporal ordering of events at the task level, Linear Temporal Logic (LTL) methods provide a correct-by-design task sequence, builds upon discretized abstractions of the robot's workspace \cite{Kress2011Correct,Belta2007Symbolic}.
This study aims to leverage incremental graph search to adapt to dynamically changing environments (as shown in Fig.~\ref{fig:spotlight}) by finding optimal navigation plans. 


While temporal-logic-based  planning offers formal guarantees on safety and provable correctness, a long-standing issue lies in its inefficiency in runtime action revision when environment states are subject to frequent, potentially unpredictable changes. A naive way is to construct a plan from scratch. However, the complexity of rewiring a feasible solution grows exponentially as robots' workspace and states scale. 
The works of \cite{Guo2013Revising}, \cite{Li2021Safe}, and \cite{Li2021Reactive} propose a local path revision based on previous plans or using behavior trees, which significantly shortens the time to replan, but they commonly lose the guarantee on optimality. Another approach is to model uncertainties \textit{a priori} and generate an offline receding-horizon strategy to react to real-time observations \cite{Kress2009Temporal,Wongpiromasarn2009Receding}.  While this approach provides an online solution, it could not offer a globally optimal solution in general.

The aforementioned approaches aim to address realizable task specifications which may become ineffective when environmental and state changes cause the tasks infeasible to achieve, e.g., to access location \textit{D} in Fig. \ref{fig:spotlight} (Top). A direct way to resolve the infeasibility is to repair the task specification or the robot's skill set \cite{Pacheck2023Automatic,Pacheck2020Finding}. To find a minimal violation motion plan that is closest to the original goal, \cite{Cai2021Learning, Guo2018Probabilistic} utilizes Markov Decision Process to acquire a policy that maximizes probabilities of satisfying given LTL task objectives. The works in \cite{Guo2015Multi} and \cite{lahijanian2016iterative} propose hard-soft constraints and enable partial violation by reclassifying specific hard constraints as soft constraints. The studies in \cite{Kim2015On} and \cite{Guo2015Multi} both present minimal violation revision strategies by relaxing product automata to remain close to the original specifications. 
In particular, \cite{Kim2015On} proposes systematically partial relaxation to find a modestly relaxed product automaton with a feasible run while similarly, \cite{Guo2015Multi} proposes a metric to quantify the task violation through atomic proposition. 
However, product automaton relaxation could result in an exceedingly large graph causing off-the-shelf algorithms to take tremendous time for a solution. Leveraging graph search algorithms for product automaton modification and relaxation offers a promising solution to address this issue.  \cite{kantaros2020stylus}, \cite{vasile2013sampling}, and \cite{vasile2020reactive} investigate efficient sampling-based approaches like RRT*, while \cite{gujarathi2022MT} and \cite{khalidi2020T}  attempt to apply heuristic-based search algorithms to solve for trajectories with guarantee on the optimality.

In this study, we will leverage the metric from \cite{Guo2015Multi} and an incremental heuristic search algorithm D* Lite \cite{Koenig2005Fast} to efficiently react to the environmental changes under both feasible and infeasible task specifications.

The main contributions of this paper are listed as:
\begin{itemize}
    \item We propose an incrementally optimal replanning algorithm for temporal-logic-based task planning problems in dynamically changing environments where the task specification is feasible to be satisfied.
    \item We propose an optimal replanning approach for the relaxed synthesized product automaton to achieve the utmost task specification satisfaction when it is infeasible to fully meet the original goal.
    \item Our replanning approach for both feasible and infeasible tasks demonstrates computational efficiency to find an optimal solution for a robot navigation problem by around two orders of magnitude.
    \end{itemize}

\section{Preliminaries}
\subsection{Linear Temporal Logic}
Linear Temporal Logic (LTL) is composed of atomic propositions $ap\in AP$ and the Boolean and temporal connectors of the syntax $\varphi := \top |\; ap \; | \;\neg \varphi \; | \;\varphi \wedge \psi \;|\;\bigcirc \varphi\;| \;\varphi\; \mathcal{U} \:\psi$, where the Boolean operators denote ``negation" ($\neg$), ``conjunction" ($\wedge$), and the temporal modalities denote ``next" ($\bigcirc$) and ``until" ($\mathcal U$). Other extended temporal connecters including ``eventually" ($\lozenge \varphi = \top \: \mathcal{U} \: \varphi$) and ``always" ($\square \varphi = \neg \lozenge \neg\varphi$) will also be used in this paper.

There exists a non-deterministic B\"{u}chi automaton (NBA) $\mathcal {A}_{\varphi}$ that can be constructed to satisfy an LTL formula $\varphi$. An NBA is defined as a tuple $\mathcal A = (Q, Q_0, 2^{AP}, \delta, \mathcal F)$ where $Q = \{q_i| i = 0,\cdots, n\}$ is a finite set of states, $Q_0 \subseteq Q$ is a set of initial states, $2^{AP}$ is a set of input alphabets, $\delta: Q \times 2^{AP} \rightarrow 2^Q$ is a transition function, and $\mathcal F \subseteq Q$ is a set of accepting states. The accepted language $\mathcal L^\omega(\mathcal A)$ of an NBA is an $\omega$-language, and consists of all infinite words $\Sigma^*$ that have a run $\sigma$ in which an accepting state is visited infinitely often. A run of such has a \emph{prefix-suffix} structure:
\begin{gather*}
\sigma = q_1 q_2 \cdots (q_f q_{f+1}\cdots q_{f+n})^\omega
\end{gather*}
where $q_1 \in Q_0$, $q_f \in \mathcal F$. The prefix of a run is executed only once, while the suffix repeats itself infinitely.

\begin{figure}
    \centering
    \vspace{.25em}
    \includegraphics[width=0.9\linewidth]{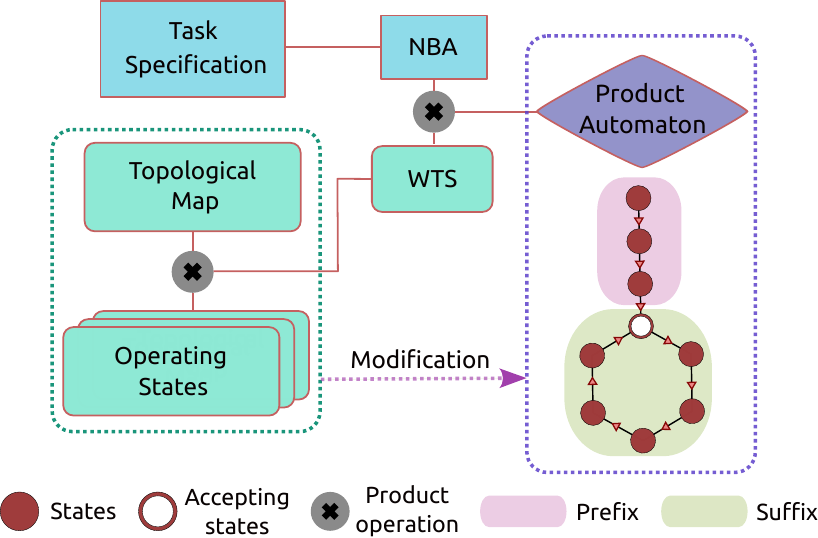}
    \caption{An illustration of the framework for synthesizing product automaton given the Weighted Transition System (WTS) and non-deterministic B\"{u}chi automaton (NBA).} 
    \label{fig:demo}
    \vspace{-0.15in}
\end{figure}

\subsection{Product Automaton}
The robot's operating states and collision-free workspace are each modeled as a weighted transition system (WTS). A WTS is a tuple $\mathcal T = (\Pi, \Pi_{\text{init}}, \rightarrow, AP, L, d)$ where $\Pi = \{\pi_i : i =0, \cdots, m\}$ is a finite set of states, $\Pi_{\text{init}} \subset \Pi$ is a set of initial states, $\rightarrow \subseteq \Pi \times \Pi$ is a transition relation in which $\pi_i \rightarrow \pi_j $ is used to express controlled transition from $\pi_i$ to $\pi_j$, $AP$ is a finite set of atomic propositions, $L : \Pi \rightarrow 2^{AP}$ is a labeling function to evaluate $ap$ to be true or false, and $d : \rightarrow \rightarrow \mathbb{R}^+$ is a positive weight assignment map for each transition, which in our study represents time consumed to relocate from one state to its successor state. More complex weight parameters can be designed for specific situations.

By encoding a robot's task planning problem as an LTL task specification $\varphi$ and by synthesizing the robot’s transition systems, we aim to generate a feasible plan on $\mathcal T$ that is at the same time accepted by $\mathcal A_{\varphi}$, given that the alphabet of $\mathcal A_{\varphi}$ consists of sets of atomic propositions in $\mathcal T$. That said, we aim to find at least one trace of WTS to be synchronously an element in $Trace(\mathcal T)$ and in the accepted language $L(\mathcal {A}_{\varphi})$. Therefore, checking that the intersection $Trace(\mathcal T) \cap L(\mathcal A_{\varphi})$ is not empty is sufficient to conclude that there is an acceptable run in the product of $\mathcal T \otimes \mathcal A_{\varphi}$. The process of the product operation is shown in Fig. \ref{fig:demo}.

The product $\mathcal T \otimes \mathcal A_{\varphi}$, named product automaton (PA), is also defined as a tuple $\Tilde{\mathcal A_{\varphi}} = (S, S_0, \delta', \mathcal F', d')$, where $S = \Pi \times Q = \{ s = \langle \pi, q \rangle | \:\forall \pi \in \Pi , \forall q\in Q \}$, $\delta': S \rightarrow 2^S$ is a transition in the condition that $\langle \pi_j, q_n\rangle \in \delta'(\langle \pi_i,q_m\rangle)$ if only if $\langle \pi_i, \pi_j\rangle \in \rightarrow$ and $q_n \in \delta(q_m, L(\pi_j))$, $S_0 = \Pi_{\rm init} \times Q_0$ is the set of initial states, $\mathcal F' = \Pi \times \mathcal F$ is the set of accepting states, and $d': \delta' \rightarrow \mathbb{R}^+$ is a cost function in the condition that $d'(\langle \pi_i,q_m\rangle, \langle \pi_j,q_n\rangle) = d(\pi_i, \pi_j)$.

\subsection{Relaxation for Product Automaton}
\label{sec: relaxtionforPA}
There are situations when a feasible run does not exist in $\Tilde{\mathcal A}_{\varphi}$ if $Trace(\mathcal T) \cap L(\mathcal A_{\varphi})=\varnothing$. To resolve this failure, one can insert a transition $\langle \pi_j, \hat q_n\rangle \in \delta'(\langle \pi_i,q_m\rangle)$ in $\Tilde{A}_{\varphi}$, where $(\pi_i, \pi_j) \in \rightarrow$ and $\hat q_n \in \delta(q_m, 2^{AP}\backslash L(\pi_i))$, to relax the initial specification $\varphi$. The work of \cite{Guo2015Multi} proposes an evaluation function that assesses the extent to which the original specification is violated from a relaxation. To quantify the violation penalty, \cite{Guo2015Multi} first designs a binary function $\xi: AP\times 2^{AP} \rightarrow \{0,1\}$ and a function $\zeta: 2^{AP} \rightarrow \{0,1\}^{|AP|}$ assuming $AP = \{ap_1, \cdots, ap_r\}$:
\begin{gather*}
 \xi(ap_i, l) = \begin{cases}
      1 & \;\text{if} \:  ap_i \in l\\
      0 & \;\text{if} \: ap_i \notin l 
    \end{cases},\;\;\;
    \zeta(l)=[\xi(ap_i, l)]^r
\end{gather*}
where $i = 1, 2 ,\cdots, r$ and $l \in 2^{AP}$. The function $\zeta$ outputs a vector of binary values denoting whether each element of $AP$ is within a subset of $AP$. A metric $ \rho: 2^{AP}\times 2^{AP} \rightarrow \mathbb{Z}^+$ is then introduced to assess the ``difference" between two subsets of $AP$:
\begin{gather*}
    \rho(l, l') = \| \zeta(l)-\zeta(l')\|_1 = \sum_{i=1}^{r}| \xi(ap_i, l) - \xi(ap_i, l')|
\end{gather*}
where $l, l' \in 2^{AP}$ and $\|\cdot\|_1$ is the $l_1$ norm. We thereby extend the definition of the ``difference" between two subsets to the ``distance" between two states, $s_m = \langle \pi_i,q_m\rangle$ and $s_n = \langle \pi_j,q_n\rangle$, in a PA, as the quantifier of violation of transition from $s_m$ to $s_n$:
\begin{gather*}
    \Call{Dist}{s_m, s_n} = \begin{cases}
      0 & \text{if} \:  l\in \chi(q_m, q_n) \\
      \min_{l'\in \chi(q_m, q_n)}\rho(l,l')  & \text{otherwise}
    \end{cases}
\end{gather*}
where $l = L(\pi_j)$ is the label of $\pi_j$ and $ \chi(q_m, q_n) = \{l\in 2^{AP}|q_n \in \delta(q_m, l)\}$ is a set containing all the subsets of $AP$ that enable the transition from $q_m$ to $q_n$.

\begin{figure}
    \centering
    \vspace{.25em}
    \includegraphics[width=0.9\linewidth]{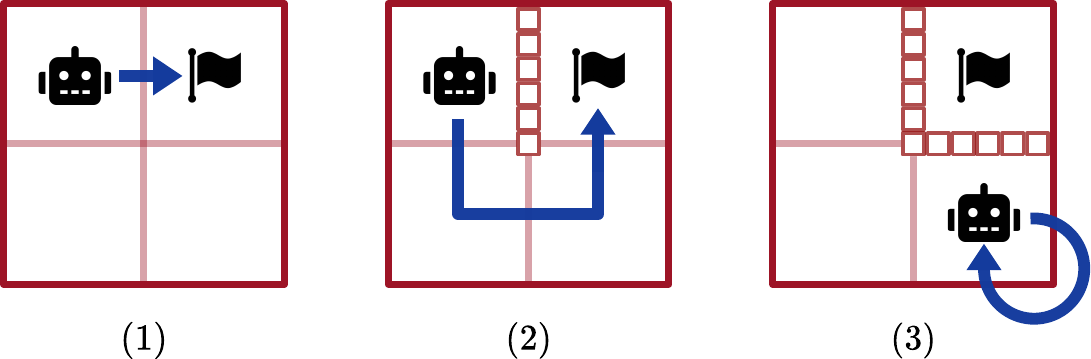}
    \caption{A conceptual illustration of feasible and infeasible tasks in our study. Assume the robot's mission is to eventually always reach the flag. In the beginning, the robot is not aware of the existence of any obstacles represented by red blocks, so it plans a direct path to the flag in (a). At runtime, it notices the obstacle in the front, so it rewires its path to the flag by a U-turn as shown in (b). This case is considered a feasible task where the task can still be fulfilled with a modified action. (c) represents a replanning in an infeasible scenario where the task is impossible to meet. When the robot reaches the bottom-right cell, it detects another obstacle, obstructing its next move. Now the robot's new plan would be to stay in the closest cell to the flag cell and maintain a minimal task violation, as it has the minimum cost in terms of traversal distance.} 
    \label{fig:demo of feasible and infeasible}
    \vspace{-0.15in}
\end{figure}

\subsection{Incremental Graph Search}
\label{sec:dstar}
Incremental graph search finds optimal solutions iteratively and is significantly faster than solving a search task from scratch. It applies to planning problems on known finite graphs whose structure evolves over time. In our study, we apply incremental search to a directed and weighted graph representing $\Tilde{\mathcal A}_{\varphi}$, denoted as $G(\Tilde{\mathcal A}_{\varphi})= (S,E)$ where $S$ is defined in $\Tilde{\mathcal A}_{\varphi}$ and $E = S\times S$ is a finite set of edges connecting a pair of states $\langle s, s'\rangle$  given that $s, s'\in S$ and $ s'\in \delta'(s)$. 
$ \Call{Cost}{s, s'}$ returns a finite value as the weight of the edge $\langle s, s'\rangle\in E$, which equals to $d'(s, s')$. 

\begin{figure}[t!]
    \centering
    \vspace{.25em}
    \includegraphics[width=0.9\linewidth]{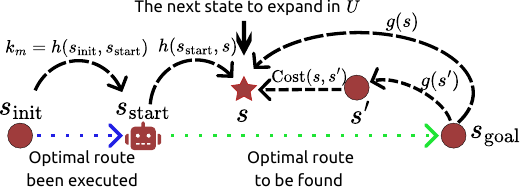}
    \caption{States in D* Lite are expanded in a reversed order from $s_{\rm goal}$ to $s_{\rm start}$ where $k_m$ is considered as the heuristic from the initial state $s_{\rm init}$ to the current robot state $s_{\rm start}$, and rhs-value of a state $s$ is updated through summation of g-value of its successor $s'$ and the edge weight of $\langle s, s' \rangle$. } 
    \label{fig:dstar explanation}
    \vspace{-0.15in}
\end{figure}

Our algorithm is based on D* Lite \cite{Koenig2005Fast} which determines the shortest path from a given start state $s_{\rm start}$ to a given goal state $s_{\rm goal}$ in $G(\Tilde{\mathcal A}_{\varphi})$. Similar to A* search, D* Lite algorithm keeps track of a priority queue $U$ and the estimates, including $g(s)$, or g-value, to denote the overall cost of the shortest path from $s\in S$ to $s_{\rm goal}$, and a heuristic $h(s)$, or h-value, to estimate cost from $s_{\rm start}$ to $s$. The heuristic function needs to be consistent to guarantee the optimality of the solution. Beyond A*, it keeps an additional estimate $rhs(s)$, or rhs-value, as a one-step lookahead value for each state defined as:
\begin{gather*}
rhs(s) = 
\begin{cases}
      0 & \text{if} \:  s \in s_{\rm goal}\\
      \min_{s'\in \Call{Succ}{s}}(g(s')+\Call{Cost}{s, s'}) & \text{otherwise}
    \end{cases}
\end{gather*}
where $\Call{Succ}{s} \subseteq S$ gives the set of successors of $s$. A state is called ``consistent" if its g-value equals to rhs-value, ``overconsistent" if its g-value is greater than the rhs-value, and ``underconsistent" if otherwise. Moreover, a duo-component key $k(s) = [\min(g(s), rhs(s))+h(s_{\rm start},s)+k_m; \;\min(g(s), rhs(s))]$ replaces f-value in A* to help decide which state to select from $U$ for the next expansion. During expansion of a state, its g-value is updated to the rhs-value if overconsistent, or set to infinity if underconsistent. $k_m$ is a heuristic modifier variable to retain the order of states in $U$ if the robot sets off from its initial state $s_{\rm init}$ to its new state $s_{\rm start}$ before a replanning is triggered, as shown in Fig. \ref{fig:dstar explanation}. We will slightly modify the design of the key $k(s)$ in our study to better suit a special construction of costs in Sec. \ref{replanningforinfeasible}.

\section {Optimal Replanning for Feasible Tasks}
\label{sec:replanningforfeasible}
In this section, we propose an on-the-fly revising algorithm that addresses unexpected environmental changes and state disturbances while the task remains feasible to achieve. We define task feasibility based on the assumption that there exists a feasible run in $\Tilde {\mathcal A}_{\varphi}$ without the need for any task specification relaxation. In the terminology of graph theory, task feasibility is equivalent to the capacity to find a path from $s_{\rm start}$ to an accepting state which has a self-referential cycle in $G(\Tilde{\mathcal A}_{\varphi})$. A conceptual example of replanning for feasible tasks can be seen in Fig. \ref{fig:demo of feasible and infeasible} (b).

The scenario considered in this section presupposes that the encoded task specifications remain consistent, with changes occurring only within the WTS. These changes in transition relations between states in $\mathcal T$ can be accordingly mapped into addition, deletion, and cost changes on edges in $G(\Tilde{\mathcal A}_{\varphi})$. The mapping operations can be defined as: assuming a transition $\langle \pi_i, \pi_j \rangle \in \rightarrow$ is added, deleted, or $d(\pi_i, \pi_j)$ is changed in $\mathcal T$, we can apply the same type of modifications between $\langle \pi_i, q_m\rangle$ and $\langle \pi_j, q_n\rangle$ in $G(\Tilde{\mathcal A}_{\varphi})$ for any $q_m, q_n$ under the condition that $q_n \in \delta\big(q_m, L(\pi_j)\big)$.

Similar to an NBA, a PA is also characterized by $\omega$-language, and an infinite run of it consists of a prefix $\sigma_{\rm pre}$ followed by an iterative suffix $\sigma_{\rm suf}$, as illustrated in Fig. \ref{fig:demo}. Even though the suffix of a run occurs infinitely often, it is impractical to ignore the cost incurred from the prefix before entering the periodic suffix loop. Additionally, it is unrealistic for a robot to run infinitely in a real-world deployment. Therefore, we design a finite parameter $\beta\in \mathbb{R}$ to represent a weighting parameter of the cost associated with the suffix, and the total cost is given as $cost_{\sigma_{\rm pre}} + \beta cost_{\sigma_{\rm suf}}$. 

\begin{figure}[t!]
    \centering
    \vspace{.25em}
    \includegraphics[width=\linewidth]{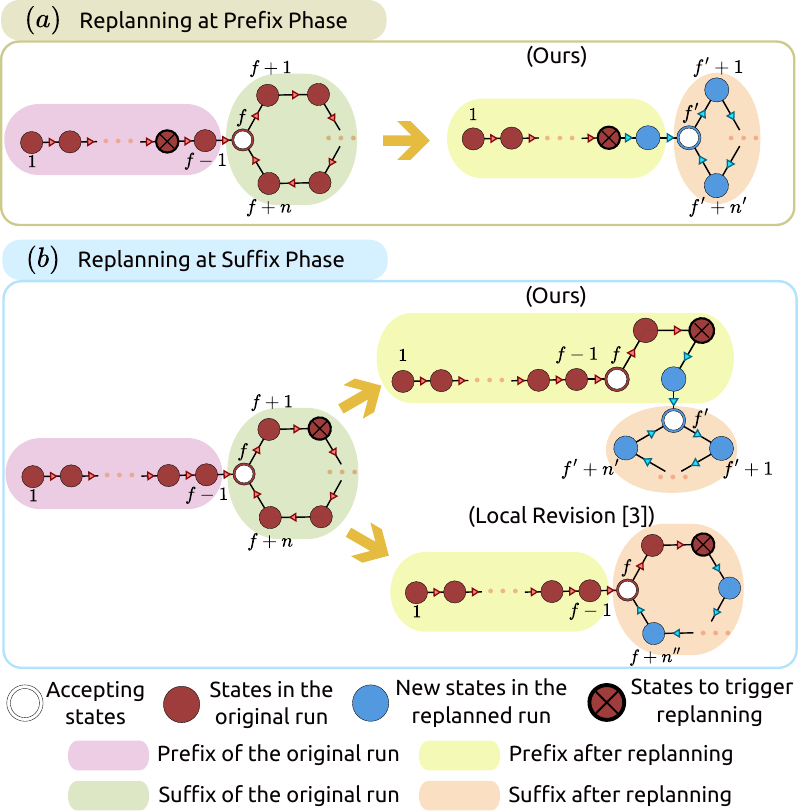}
    \caption{An illustration for replanning strategies when modification is performed in the prefix or suffix phase of a run. This adaptation is caused by the modification within WTS, and corresponding edge changes in PA impact the optimality of the original run. Note that, $f$ and $f'$ denote the index of the accepting states, and $n$, $n'$ and $n''$ denote the index of the last element of the suffixes.} 
    \label{fig:prefixandsuffix}
    \vspace{-0.15in}
\end{figure}

\subsection{Revision for Prefix and Suffix}
\label{subsec:revision for prefix and suffix}

In this subsection, we propose two distinct replanning strategies when a robot is progressing through different phases: (i) if the robot is currently in the prefix phase (as shown in Fig.~\ref{fig:prefixandsuffix}(a)), we simply find a new accepting state that gives the minimal total cost of the trace guiding the robot from the current state to the new accepting state and then repeat the minimal-cost loop of the new accepting state; (ii) if the robot is currently in the suffix phase (as illustrated in Fig. \ref{fig:prefixandsuffix}(b)), the optimal solution will first maintain the preceding trajectory ahead of the current state as part of the new prefix, and then find a new accepting state that renders the minimal total cost, similar to that in case (i). 

We show that our replanning strategies outperform the local revision method \cite{Guo2013Revising} in term of a lower cost of the solution. The local revision algorithm identifies the shortest detour upon its prior plan and ultimately guides it back to the original route. This replanning strategy independently rewires the prefix and suffix segments of a feasible path. 

\begin{algorithm}[t!]
	\caption{$\protect \Call{SuffixInitialize}{}$}
        \label{alg:SuffixInitialize}
        \hspace*{\algorithmicindent} \textbf{Input}: A graph $G(\Tilde{ \mathcal A}_\varphi)$, index $k$\\
        \hspace*{\algorithmicindent} \textbf{Output}: The optimal suffix $\sigma_{\rm suf}^{\rm k}$, and its cost $cost[s_{\rm acc}^{\rm k}]$
	\begin{algorithmic}[1]
            \State $\Call{Initialize}{\null}$ with superscript $k$ on all variables.
            \State Construct $s^{\rm k}_{\rm img}$.
            \For {$s' \in \Call{Pred} {s^{\rm k}_{\rm acc}}$}
            \State Insert $\langle s',s^{\rm k}_{\rm img}\rangle$ with the weight $\Call{Cost}{s',s^{\rm k}_{\rm acc}}$.
            \EndFor
            \State $\Call{ComputeShortestPath}{s_{\rm acc}^{\rm k}, s^{\rm k}_{\rm img}}$
            \State Retrieve $\sigma_{\rm suf}^{\rm k}$ and the intergal loop cost $cost[s_{\rm acc}^{\rm k}]$.
	\end{algorithmic} 
\end{algorithm} 

\begin{algorithm}[t!]
	\caption{$\protect \Call{SuffixReplan}{}$}
        \label{alg:SuffixReplan}
        \hspace*{\algorithmicindent} \textbf{Input}:  A set of modified edges $mod$, index $k$ \\
        \hspace*{\algorithmicindent} \textbf{Output}: The optimal suffix $\sigma_{\rm suf}^{\rm k}$, and its cost $cost[s_{\rm acc}^{\rm k}]$
	\begin{algorithmic}[1]
            \For { $\langle u,v \rangle \in mod$}
            \If {$v$ = $s^{\rm k}_{\rm acc}$}
            \State Update the weight $\Call{Cost}{u,s^{\rm k}_{\rm img}}$
            \EndIf
            \State Update the weight $\Call{Cost}{u,v}$
            \State $\Call{UpdateVertex}{u}$
            \EndFor
            \State $\Call{ComputeShortestPath}{s_{\rm acc}^{\rm k}, s^{\rm k}_{\rm img}}$
            \State Retrieve $\sigma_{\rm suf}^{\rm k}$ and the intergal loop cost $cost[s_{\rm acc}^{\rm k}]$.
	\end{algorithmic} 
\end{algorithm}


To argue that our strategy gives a lower cost, we assign a substantially large value to $\beta$ to emphasize the major contribution of the suffix due to its repetitive occurrence in a run. For the local revision method shown in Fig.~\ref{fig:prefixandsuffix}(b), the overall cost $cost_{\sigma'}$ is:
\begin{gather*}
cost_{\sigma'} =
\sum_{i=1}^{f-1}\Call{Cost}{s_i, s_{i+1}} + \beta\sum_{i=f}^{f+n''}\Call{Cost}{s_i. s_{i+1}}
\end{gather*}
where $f$ is the index of the original accepting state, and $n''$ is the index of the last element in the suffix. Note that, we define $s_{f+n''+1} = s_{f}$. Similarly, the overall cost $cost_{\sigma}$ using our strategy to find a new accepting state and its minimal-cost suffix loop is:
\setlength{\jot}{-6pt}
\begin{equation*}
\begin{gathered}
cost_{\sigma} =
\sum_{i=1}^{f-1}\Call{Cost}{s_i, s_{i+1}} + \sum_{i=f}^{f'-1}\Call{Cost}{s_i, s_{i+1}} \\ + \beta\sum_{i=f'}^{f' + n'}\Call{Cost}{s_i, s_{i+1}}
\end{gathered}
\end{equation*}
\setlength{\jot}{0pt}
where $f'$ is the index of the new accepting state, and $n'$ is the index of the new last element in the suffix. We again define $s_{f'+n'+1} = s_{f'}$.
It is intuitive to show that $\beta\big(\sum_{i=f}^{f+n''}\Call{Cost}{s_i. s_{i+1}}- \sum_{i=f'}^{f'+n'}\Call{Cost}{s_i, s_{i+1}}\big) > \sum_{i=f}^{f'-1}\Call{Cost}{s_i, s_{i+1}} $ when $\beta$ is substantially large, indicating $cost_{\sigma'}>cost_{\sigma}$, and therefore, our strategy outperforms in term of a lower cost.


\begin{algorithm}[t!]
	\caption{LTL-D*}
        \label{alg:LTL-D*}
	\begin{algorithmic}[1]
            \State Construct $G(\Tilde{ \mathcal A}_\varphi)$ and $\Call{Initialize}{\null}$.
            \State Construct $s_{\rm img}$.
            \For {$s^{\rm k}_{\rm acc} \in \mathcal F'$}
            \State $\sigma_{\rm suf}^{\rm k}, cost[s_{\rm acc}^{\rm k}] = \Call{SuffixInitialize}{G(\Tilde{ \mathcal A}_\varphi), k}$
            \State Insert $\langle s_{\rm acc}^{\rm k},s_{\rm img}\rangle$ with the weight $cost[s_{\rm acc}^{\rm k}]$.
            \EndFor
            \State $\Call{ComputeShortestPath}{s_{\rm start}, s_{\rm img}}$
            \State Retrieve $\sigma_{\rm pre}$ and $\sigma_{\rm suf}$.
            \While {True}
            \State Move to the next state $s_{\rm next}$. Let $s_{\rm start} = s_{\rm next}$.
            \State Scan graph for a set $mod$ containing changed edges.
            \If {$mod$}
            \For {$s^{\rm k}_{\rm acc} \in \mathcal F'$}
            \State \begin{varwidth}[t]{\linewidth} $\sigma_{\rm suf}^{\rm k}, cost[s_{\rm acc}^{\rm k}] =\Call{SuffixReplan}{mod, k}$ \end{varwidth}
            \If {$cost[s_{\rm acc}^{\rm k}]$ is updated}
            \State $\Call{UpdateVertex}{s_{\rm acc}^{\rm k}}$ 
            \EndIf
            \EndFor
            \State Update $k_m$ \textbf{if} in $\sigma_{\rm pre}$ \textbf{else} $\Call{Initialize}{\null}$
            \For {$\langle u, v \rangle \in mod$}
            \State Update the weight with $\Call{Cost}{u,v}$.
            \State $\Call{UpdateVertex}{u}$
            \EndFor
            \State $\Call{ComputeShortestPath}{s_{\rm start}, s_{\rm img}}$
            \State Retreive $\sigma_{\rm pre}$ and $\sigma_{\rm suf}$.
            \EndIf
            \EndWhile
	\end{algorithmic} 
\end{algorithm} 

Regardless of the current phase (i.e., either prefix or suffix) that the robot stays at, replanning using our strategy needs to construct both new prefix and suffix segments. To do so, we first find the minimal-cost loop $\sigma_{\rm suf}^{\rm k}$ in $G(\Tilde{\mathcal A}_{\varphi})$ starting and ending at each accepting state $s^{\rm k}_{\rm acc} \in \mathcal F'$ where the index $k =1,2,\cdots,w$ and $w$ is the total number of accepting states. More details will be provided in Sec.~\ref{subsec:find-suffix}. Following that, we find the shortest path from the current state to the accepting state that renders the minimal total cost, given the knowledge of the smallest suffix cost $cost[s_{\rm acc}^{\rm k}]$ each accepting state possesses. More details for this step will be provided in Sec.~\ref{subsec:find-prefix}. A psudocode implementation is outlined in Algorithm~\ref{alg:LTL-D*} which is designed based on the original D* Lite algorithm.\footnote{Algorithm~\ref{alg:LTL-D*} calls a series of functions $\Call{ComputeShortestPath}{}$, $\Call{UpdateVertex}{}$, $\Call{Initialize}{}$ from the original D* Lite algorithm \cite{Koenig2005Fast}, as well as $\Call{SuffixInitialize}{}$ and $\Call{SuffixReplan}{}$ in Algorithms \ref{alg:SuffixInitialize} and \ref{alg:SuffixReplan}.} The following two subsections will provide detailed explanations on Algorithms \ref{alg:SuffixInitialize}, \ref{alg:SuffixReplan} and \ref{alg:LTL-D*}. 


\subsection{Searching for Optimal Suffix Loops}
\label{subsec:find-suffix}
\begin{figure}[t!]
    \centering
    \vspace{.25em}
    \includegraphics[width=0.9\linewidth]{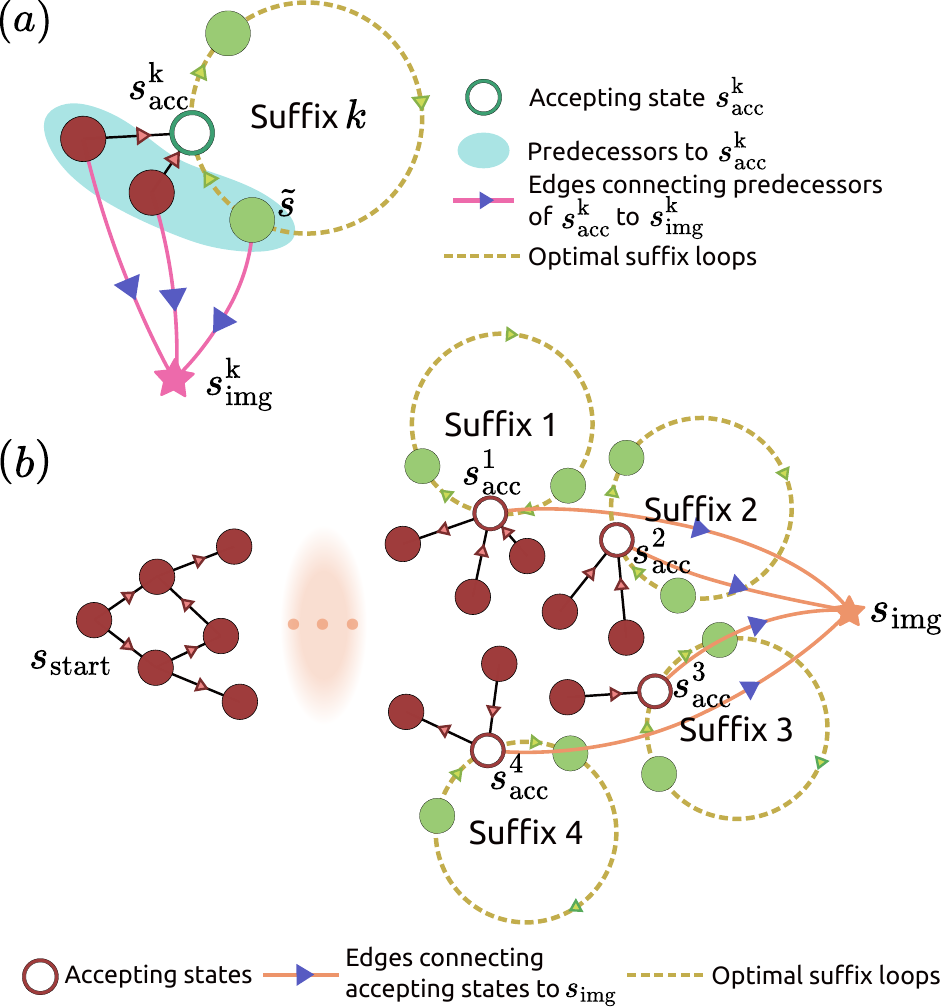}
    \caption{In (a), we find an optimal suffix loop starting and ending at the accepting state $s^{\rm k}_{\rm acc}$. We introduce $s^{\rm k}_{\rm img}$ and connect all predecessors of $s^{\rm k}_{\rm acc}$ to  $s^{\rm k}_{\rm img}$. In (b), we find the shortest path from $s_{\rm start}$ to the accepting state with the minimal total cost. We introduce an imaginary goal $s_{\rm img}$ and connect all accepting states to $s_{\rm img}$. We assume $w=4$ in this illustration.} 
    \label{fig:prefix}
    \vspace{-0.15in}
\end{figure}

This subsection focuses on finding an optimal suffix $\sigma^{\rm k}_{\rm pre}$ in $G(\Tilde{\mathcal A}_{\varphi})$ starting and ending at $s^{\rm k}_{\rm acc}$. An illustration of this process is shown in Fig. \ref{fig:prefix} (a). Particularly, we leverage the concept of the imaginary goal to serve as a single virtual target node for the search algorithm. First, we construct an imaginary goal $s_{\rm img}^{\rm k}$ (\textit{Algorithm 1 Line 2}) and connect all predecessors of $s_{\rm acc}^{\rm k}$ to $s_{\rm img}^{\rm k}$ (\textit{Alogorithm 2 Lines 3-4}). Suppose that there does exist a cyclic path originating and terminating at $s^{\rm k}_{\rm acc}$ (otherwise we simply set $cost[s_{\rm acc}^{\rm k}] = \infty$). By computing a shortest path from $s_{\rm acc}^{\rm k}$ to $s_{\rm img}^{\rm k}$ and generating a course of $s_{\rm acc}^{\rm k}\cdots \tilde{s} \:s_{\rm img}^{\rm k}$ as ouput,  where $\tilde{s} \in \Call{Pred}{s_{\rm acc}^{\rm k}}$, we can project the course to be $s_{\rm acc}^{\rm k}\cdots \tilde{s} \:s_{\rm acc}^{\rm k}$ and regard it as the optimal suffix of $s_{\rm acc}^{\rm k}$ (\textit{Algorithm 1 Lines 5-6}). Whenever the operations (i.e., addition, deletion or cost changes) are applied to the edges connecting a predecessor of $s^{\rm k}_{\rm acc}$ to $s^{\rm k}_{\rm acc}$, we also map the operations onto the edge leading from the same predecessor to $s^{\rm k}_{\rm img}$ that mirrors the accepting state (\textit{Algorithm 2 Lines 2-3}). 

\subsection{Searching for Optimal Total Path}
\label{subsec:find-prefix}
After iterating through all accepting states and identifying the smallest suffix cost for each, we will determine the shortest path from the current state to the accepting state with the minimal total cost. However, a run to any accepting state that has a suffix cost not equal to infinity could form a valid run. Therefore, similar to the approach in Sec.~\ref{subsec:find-suffix}, we append an imaginary goal $s_{\rm img}$ to $G(\Tilde{\mathcal A}_{\varphi})$, and build edges connecting all accepting states to $s_{\rm img}$ with the associated cost $cost[s_{\rm acc}^{\rm k}]$ depending on the accepting state index (\textit{Algorithm 3 Lines 3-5}), as shown in Fig. \ref{fig:prefix} (b). We then compute a shortest route from the current state to $s_{\rm img}$, in the sequence of $s_{\rm start} \cdots s_{\rm acc}^{\rm i}\: s_{\rm img}$, where the accepting state $s_{\rm acc}^{\rm i}$ renders the minimal total cost (\textit{Algorithm 3 Line 4}). This path in $G(\Tilde{\mathcal A}_{\varphi})$ corresponds to the optimal run $s_{\rm start} \cdots \{s_{\rm acc}^{\rm i} \cdots \}^{\omega}$ in $\Tilde{\mathcal A}_{\varphi}$, where $\{s_{\rm acc}^{\rm i} \cdots\}$ is the minimal-cost loop starting from $s_{\rm acc}^{\rm i}$ found in Sec. \ref{subsec:find-suffix} (\textit{Algorithm 3 Line 5}). Every time a set of edges, denoted as $mod$, in $G(\Tilde{\mathcal A}_{\varphi})$, undergoes changes (\textit{Algorithm 3, Line 11}), we first call $\Call{SuffixReplan}{}$ to retrieve new suffix costs (\textit{Algorithm 3, Lines 12-13}). If $cost[s_{\rm acc}^{\rm k}]$ is updated after replanning, we count $\langle s_{\rm acc}^{\rm k},  s_{\rm img} \rangle$ as a modified edge and update the rhs-value and key of $s_{\rm acc}^{\rm k}$ through $\Call{UpdateVertex}{}$ (\textit{Algorithm 3, Lines 14-15}). 

\section {Optimal Replanning for Infeasible Tasks}
\label{replanningforinfeasible}
The revision algorithm discussed in the preceding section only addresses replanning scenarios when any modification to the weighted transition system $\mathcal T$ can generate a feasible run in $\Tilde{\mathcal{A_{\varphi}}}$. Nevertheless, there exist cases in which environmental changes or state disturbances may induce an infeasible run, e.g., the example shown in Fig. \ref{fig:demo of feasible and infeasible} (c). In such cases, any potential run whose projection onto the WTS $\mathcal T$ violates the original task specification to a certain degree. To address this challenge, we further optimize the proposed algorithm in Sec. \ref{sec:replanningforfeasible} by leveraging the distance metric introduced in Sec. \ref{sec: relaxtionforPA}. Our approach aims to identify an optimal run that minimally deviates from the original task specification while incurring the lowest cost, thereby guaranteeing optimality. Despite this new priority for minimal task violation, we will demonstrate that our method still maintains the same efficiency as the algorithm in Sec. \ref{sec:replanningforfeasible}.


Given the updated priority, we assign the weight to the relaxed edge $\langle s, s'\rangle$ as a combination of the transition cost $\Call{Cost}{s, s'}$ and the violation penalty $\Call{Dist}{s, s'}$. To integrate this combinatory cost to our incremental search algorithm, we introduce two auxiliary estimates, $g_{\rm aux}(s)$ and $rhs_{\rm aux}(s)$, alongside the original g-value and rhs-value in D* Lite. These auxiliary estimates respectively serve as a real estimate and one-step lookahead to the accumulation of the task violation penalty given by $\Call{Dist}{}$ and both are initialized infinity. Since there are two metrics to consider, we therefore revise the original duo-component key in the priority queue as:
\begin{gather*}
\begin{aligned}
k(s) = \{& \min\big(\overline {g \vphantom{h}}(s), \overline {rhs}(s)\big) + h(s_{\rm start}, s)+k_m; \\ & \min\big(\overline {g \vphantom{h}}(s), \overline {rhs}(s)\big)\}
\end{aligned}
\end{gather*}
We define $\overline {g \vphantom{h}}(s)$ and $\overline {rhs}(s)$ as the combinations of violation penalties and other costs, given by
\begin{gather*}
\begin{aligned}
\overline {g \vphantom{h}}(s) & = g(s) + \gamma g_{\rm aux}(s), \\
\overline {rhs}(s) &= \underbrace{g(s') + \Call{Cost}{s,s'}}_{rhs(s)} + \gamma\: \underbrace{\big(g_{\rm aux}(s') +\Call{Dist}{s, s'}\big)}_{rhs_{\rm aux}(s)}
\end{aligned}
\end{gather*}
where $s'$ is among the successors of $s$:
\setlength{\jot}{-6pt}
\begin{multline*}
s' = \argmin_{s \in \Call{Pred}{u}} \Big( g(s) + \Call{Cost}{s,u} \\ + \gamma \big(g_{\rm aux}(s) +\Call{Dist}{s, v}\big)\Big)
\end{multline*}
\setlength{\jot}{0pt}
and we assume a constant $\gamma \gg g(s)$ so that all auxiliary estimates are dominant in determining the weight of each component in the key during ranking, i.e., if $g_{\rm aux}(s_1) > g_{\rm aux}(s_2)$, then we have $\overline {g \vphantom{h}}(s_1) > \overline {g \vphantom{h}}(s_2)$. Since keys are compared according to lexicographic ordering in $U$, this modified design prioritizes the minimal violation over other costs. 

We contend that our approach to revising the key design coheres with \cite{koenig2002improved} regarding how a state is expanded and how a shortest path is retrieved through backtracking. Specifically, we argue that the following propositions stand true: \begin{itemize}
    \item \textit{State consistency infers consistency of both the violation penalty and the total cost}: Recall that in Sec.\ref{sec:dstar} we define state consistency as $\overline {g \vphantom{h}}(s) = \overline{rhs}(s)$. If state $s$ is consistent, we can derive that $g(s) = rhs(s)$ and $g_{\rm aux}(s) = rhs_{\rm aux}(s)$. This can be shown by contradiction. \footnote{If $g_{\rm aux}(s) \neq rhs_{\rm aux}(s)$, because of the weighting $\gamma \gg g(s)$,  $\overline {g \vphantom{h}}(s) \neq \overline{rhs}(s)$. If $g_{\rm aux}(s) = rhs_{\rm aux}(s)$ but $g(s)\neq rhs(s)$, still $\overline {g \vphantom{h}}(s) \neq \overline{rhs}(s)$.}
    \item \textit{Consistency of the heuristic function still holds}: This is proved by arguing that we use the same admissible heuristic function for costs, and after the violation penalty being added, we have $h(s_{\rm start},s) \leqslant h(s_{\rm start},s') + \Call{Cost}{s',s} \leqslant  h(s_{\rm start},s') + \Call{Cost}{s',s} + \gamma\Call{Dist}{s,s'})$. Therefore, consistency of the heuristic function still holds.
    \item \textit{The shortest path has the least violation of task specification and the lowest traveling time}: A shortest path is backtracked by always moving forward from the current state $s$ to any successor $s'$ that minimizes $\overline {g\vphantom{h}}(s') +  \Call{Cost}{s,s'} + \gamma\Call{Dist}{s,s'}$ until $s_{\rm goal}$ is reached. This will ensure that the state for the next move is the state with the lowest $g_{\rm aux}$ among all successors of $s$, and if multiple states own the same $g_{\rm aux}$ value, then it has the lowest $g$ value.
\end{itemize}
Indeed, with all the propositions stated above, the correctness and optimality of our algorithm can be proved in a similar way as the lemmas and theorems proved in \cite{koenig2002improved}. The proofs have to be omitted due to limited space. 
Moreover, modeling the key in this way provides efficiency in finding an optimal path when the robot lacks prior knowledge of the task's feasibility. If there exists a non-violating run, states that are not affected by LTL specification relaxation will be expanded and their keys will turn consistent before any state impacted by relaxation is expanded. This is because those states not affected by relaxation has $\min(g_{\rm aux}, rhs_{\rm aux})=0$,  which leads to their keys ranking higher in the priority queue than those affected by relaxation.

\section {Results}
\subsection{Benchmarking Environment and Task Specifications}
To demonstrate the efficiency and optimality of our proposed algorithm, we test on the benchmark maps where the robot only has partial observation on its local environment shown in Fig.~\ref{fig:result1} (a) and (b). The robot's task is to repetitively visit locations \textit{A}, \textit{B}, \textit{C}, \textit{D} sequentially, expressed in LTL formula as follows:
\begin{gather*}
\begin{aligned}\label{eq:task-spec1}
\varphi_b &= \square(A \rightarrow \bigcirc\big((\neg A \wedge \neg D \wedge \neg C) \:\mathcal U\: \varphi_1\big)
\end{aligned}
\end{gather*}
where $\varphi_1 = B \wedge \bigcirc\big((\neg B \wedge \neg A \: \wedge \neg D) \; \mathcal U \: \varphi_2\big)$, $\varphi_2 = C \wedge \bigcirc\big((\neg C \wedge \neg B \wedge \neg A) \;\mathcal U \: \varphi_3\big)$, and $\varphi_3 = D \wedge \bigcirc\big((\neg D  \wedge \neg C \wedge \neg B) \;\mathcal U \: A\big)$. The map is configured as an N-by-N gridworld, which has four adjacent sub-regions comprising A, B, C, and D locations respectively. This map includes static obstacles and walls that impede the robot's movement, as well as static bumps that cause delays in travel time. Weighted transition system weights are assigned based on traversal time between cells, with a default cost of 10 for regular movements and an increased cost of 50 for traversing bumps. Initially, the robot possesses only partial knowledge of the map, being informed solely about the map size and the positions of walls. No information is provided regarding the locations of obstacles and bumps until the robot enters cells adjacent to these objects, i.e., one-cell horizon for the robot. The robot needs to dynamically respond to obstacles blocking its path and bumps slowing down its speed, and find an alternative optimal path as needed. 

\subsection{Baseline Benchmarking}
In replanning scenarios where the tasks are feasible, we compare two baseline algorithms: the local revision method introduced in Sec.~\ref{subsec:revision for prefix and suffix} and iterative planning. The iterative planning algorithm simply replans from the current state using Dijkstra's algorithm whenever its next planned action is blocked or impeded by an object. We assign $\beta=10$ as the relative weighting when calculating the integral cost.
Our LTL-D* method
uses Manhattan distance as the heuristic function to guide the search and $\beta$ is also assigned as 10 to align with the relative weighting between costs of prefix and suffix as the same as that in the iterative planning.

\begin{figure}
    \centering
    \vspace{.25em}
    \includegraphics[width=\linewidth]{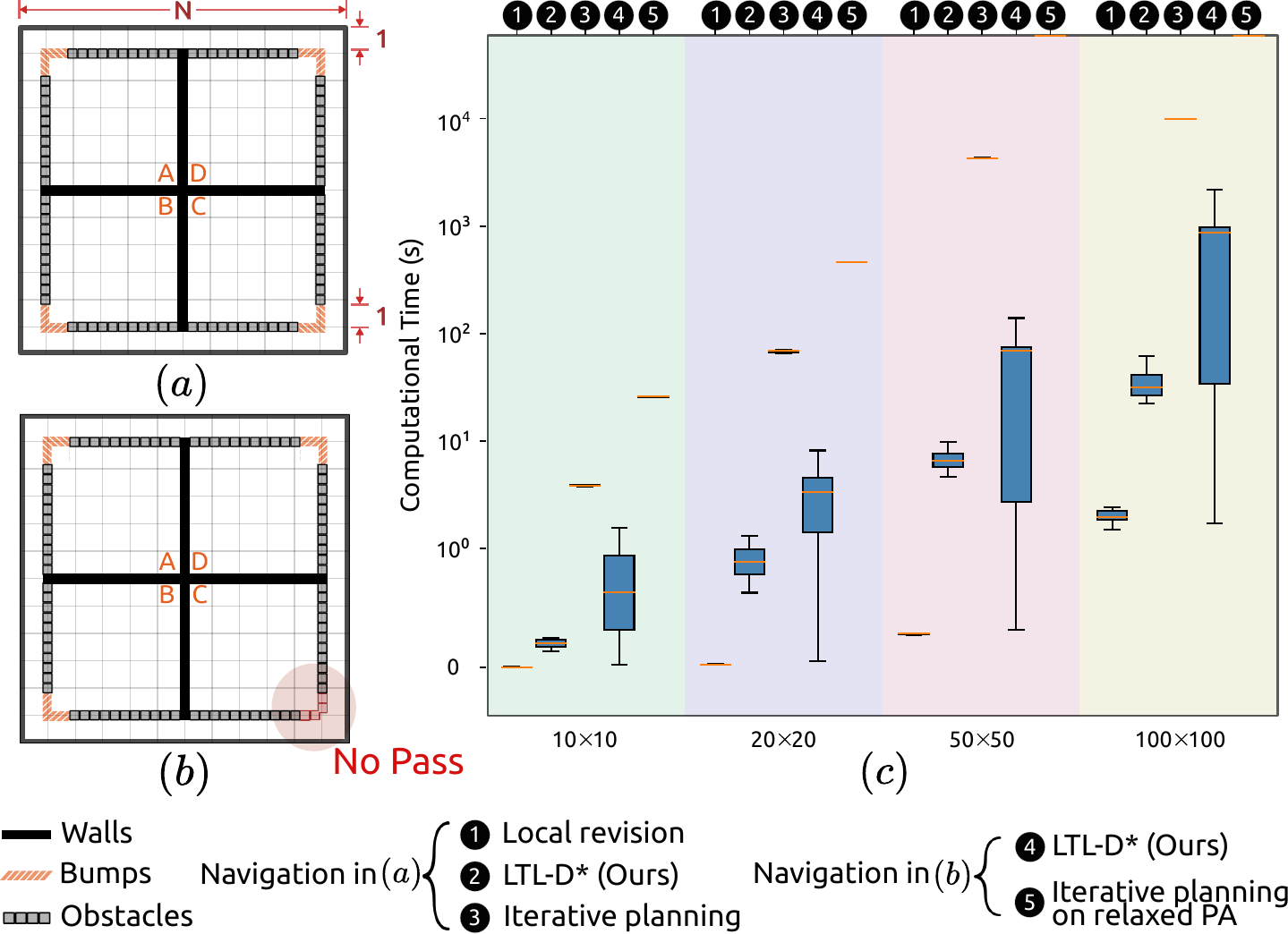}
    \caption{The comparison of the computational time for replanning every time the robot encounters new obstacles and bumps between the baseline algorithms and our approach in benchmark map (a) where $\varphi_b$ is feasible to realize and in benchmark map (b) where $\varphi_b$ is infeasible to realize because red obstacles block the passage to reach location \textit{C}. The computational time that exceeds the y-axis limit is positioned on the frame of (c). The local revision method does not guarantee the optimality of the revised solution (shown in Fig. \ref{fig:result2}) even though it employs local adjustments more efficiently.} 
    \label{fig:result1}
    \vspace{-0.05in}
\end{figure}

\begin{table}[t!]
\begin{tabular}{p{6mm}p{10mm}p{10mm}p{10mm}p{12mm}p{12mm}}
\multicolumn{2}{c}{\backslashbox[25mm]{Tuple}{Map\\Size}} & $10 \times 10$  & $20 \times 20$   &  $50 \times 50$  & $100 \times 100$  \\ \hline  
NBA & & \multicolumn{4}{c}{32 states, 92 transitions}  \\    \hline     
WTS & states \newline transitions & 100  \newline 428  & 400  \newline 1848 & 2500 \newline 12108 & 10000 \newline 49008 \\ \hline
PA & states \newline transitions & 3200  \newline 20058  & 12800  \newline 88218 & 80000 \newline 580698 & 320000 \newline 2361498 \\ \hline
relaxed-PA & states \newline transitions & 3200  \newline 39376  & 12800  \newline 170016 & 80000 \newline 1113936 & 320000 \newline 4527136 \\ \hline
\end{tabular}
\caption {\label{tab:comparison} Number of states and transitions in the NBA, WTS, PA, and relaxed-PA for different sizes of maps.} 
\end{table}

For fair comparisons, we implement all algorithms in Python and run on a computer with a 13th Gen Intel Core i9-13900K CPU. We also avoid the usage of any pre-implemented algorithms from off-the-shelf packages. The algorithms are tested on the benchmarking map with the gridworld size of $N =10, 20, 50, 100$.

\begin{figure}
    \centering
    \vspace{.25em}
    \includegraphics[width=\linewidth]{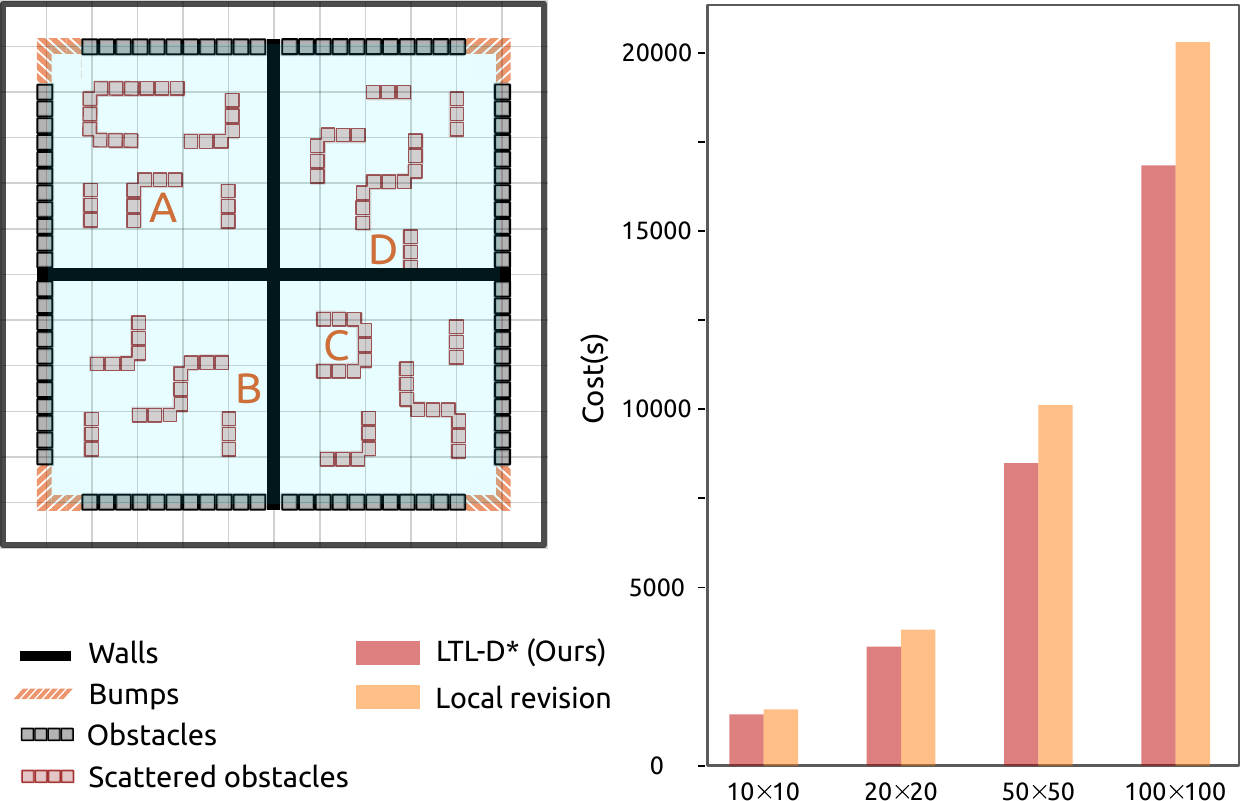}
    \caption{The average total cost to finish one loop to and from \textit{A} out of three trials for each map size using the local revision method and our method. The map is scattered with obstacles and the locations of \textit{A}, \textit{B}, \textit{C}, \textit{D} are picked randomly in the four quadrants where they belong.} 
    \label{fig:result2}
    \vspace{-0.15in}
\end{figure}

Tab. \ref{tab:comparison} shows the number of states and transitions in NBA, WTS, and PA for each map size in Fig. \ref{fig:result1}(a) and (b), and Fig. \ref{fig:result1}(c) shows the lower bound, upper bound, median, and first quartile to the third quartile of computational time data whenever a replanning is triggered, i.e., encountering unknown obstacles or bumps in its next move. In a similar environment setting in Fig \ref{fig:result2} with scattered obstacles in the blue region at the density of 40\%, we calculate the total time for the agent to finish a round trip through $A$, $B$, $C$, $D$ and returns to $A$. Our algorithm showcases a remarkable reduction in computational time compared to iterative planning, achieving speeds two orders of magnitude faster. While it still lags behind local revision in computational time by a factor of ten, our algorithm increasingly outperforms local revision in total cost to finish one run as the map size scales, as shown in Fig. \ref{fig:result2}.

For replanning for infeasible tasks, we modify our benchmark map in Fig. \ref{fig:result1} (b) by blocking the only passage that existed at the bottom right corner to access point \textit{C}. Now whatever action the robot takes, the task specification $\varphi_b$ cannot be satisfied, and the robot needs to go around to \textit{D} directly after reaching \textit{B}. We again employ iterative replanning as a baseline algorithm but this time to wire a route in a fully relaxed product automaton which is constructed \textit{a priori} under the relaxation condition in Sec. \ref{sec: relaxtionforPA}. We demonstrate a more than two-order of magnitude increase in speed to find an optimal run with the least violation and the associated minimal cost of our method compared to iterative planning in the fully relaxed PA.  

\subsection{Simulation Results}
\begin{figure}
    \centering
    \vspace{.25em}
    \includegraphics[width=0.8\linewidth]{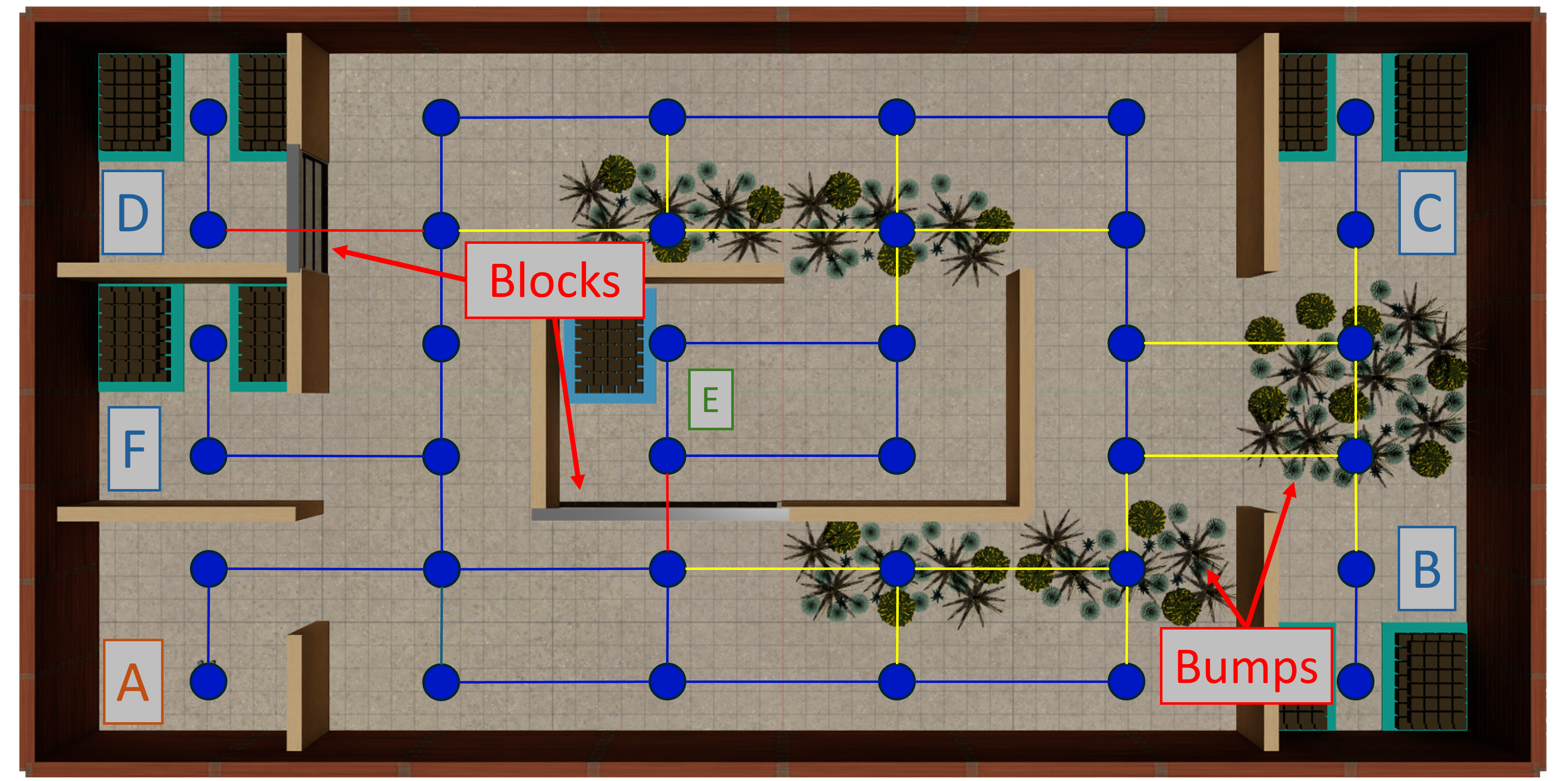}
    \caption{The simulation environment is encapsulated by a set of waypoints connecting the different rooms. Blue connections describe a normal pathway, yellow connections describe a connection that passes through a bump, and red connections describe a block that prevents passage. Connections do not exist through walls as these are always impassable.} 
    \label{fig:simulation-diagram}
    \vspace{-0.15in}
\end{figure}

We demonstrate the feasibility of our proposed algorithm by a drone navigation problem in a more realistic environment built in NVIDIA Isaac Sim. A maze-like environment is constructed as shown in Fig. \ref{fig:simulation-diagram} with a charging station \textit{A}, four rooms with cargo \textit{B}, \textit{C}, \textit{D}, \textit{F}, and a central drop-off location \textit{E}. We designate the bumps as foliage in the environment that the drone can traverse with a greater cost, and blocks as closed doors that the drone cannot traverse through. A 6-by-6 grid is overlayed in the environment to provide the robot with a set of waypoints that encompass the key locations and possible paths. For the task specifications, we require the drone to start at charging room \textit{A}. It is then instructed to load boxes from one of the rooms \textit{B}, \textit{C}, \textit{D}, \textit{F}, and bring the loaded boxes one at a time to the central drop-off location \textit{E}. The drone will then go to another room to load and unload boxes until all boxes have been delivered from each room to \textit{E}. In this scenario, only the locations of walls for each room are known while all other blocks and bumps are unknown to the robot. We can express the LTL specification for this problem as:
\begin{gather*}
\begin{aligned}
    \varphi_1 = &(A \rightarrow \lozenge B) \wedge \square (B \rightarrow \bigcirc (loaded \wedge \\
                &(\neg C \wedge \neg D \wedge \neg F) \mathcal U (E \wedge unloaded))) \\ 
    \varphi_2 = &(A \rightarrow \lozenge C) \wedge \square (C \rightarrow \bigcirc (loaded \wedge \\
                &(\neg B \wedge \neg D \wedge \neg F) \mathcal U (E \wedge unloaded))) \\ 
    \varphi_3 = &(A \rightarrow \lozenge D) \wedge \square (D \rightarrow \bigcirc (loaded \wedge \\
                &(\neg B \wedge \neg C \wedge \neg F) \mathcal U (E \wedge unloaded))) \\ 
    \varphi_4 = &(A \rightarrow \lozenge F) \wedge \square (F \rightarrow \bigcirc (loaded \wedge \\
                &(\neg B \wedge \neg C \wedge \neg D) \mathcal U (E \wedge unloaded))) \\ 
\end{aligned}
\end{gather*}
where each specification describes the robot going to a room, picking up its load, and delivering it to the drop-off location. The overall task specification can be expressed as:
\setlength{\jot}{-6pt}
\begin{gather*}
    \varphi_s = \varphi_1 \wedge \varphi_2 \wedge \varphi_3 \wedge \varphi_4
\end{gather*}
\setlength{\jot}{0pt}
Fig. \ref{fig:spotlight} (Top) shows the drone's trajectory to complete the given task specification. The yellow markers denote locations where $\varphi_s$ remains feasible after encountering an unexpected object in the environment. An infeasible task replanning, denoted by a red marker, is triggered when the robot is trying to load a box from \textit{D} which is obstructed by a closed door. The robot remains at its position thereafter as no further actions is instructed.

\section {Conclusion}


In this paper, we propose an optimal incremental replanning strategy for both feasible and infeasible task specifications. We demonstrate the efficiency, optimality, and scalability of our algorithm in benchmark maps and its application in a realistic scenario. 
Future work includes extending our algorithms to efficient and optimal task reallocation and motion replanning in multi-agent coordination.
\bibliographystyle{IEEEtran}
\bibliography{references.bib}

\end{document}